\documentclass{article}

\usepackage[final]{nips_2016}

\usepackage[utf8]{inputenc} 
\usepackage[T1]{fontenc}    
\usepackage{hyperref}       
\usepackage{url}            
\usepackage{booktabs}       
\usepackage{amsfonts}       
\usepackage{nicefrac}       
\usepackage{microtype}      
\usepackage{natbib}
 
\usepackage[pdftex]{graphicx}
\graphicspath{{./images/}}
\DeclareGraphicsExtensions{.pdf,.jpg,.png}

\title{Incorporating the Knowledge of Dermatologists to Convolutional Neural Networks for the Diagnosis of Skin Lesions}

\author{
  Iv\'an Gonz\'alez-D\'iaz \\
  Department of Signal Theory and Communications\\
  Universidad Carlos III de Madrid\\
  Legan\'es, 28911, Spain\\
  \texttt{igonzalez@tsc.uc3m.es} \\
}

\begin{document}

\maketitle

\begin{abstract}
  This report describes our submission to the \emph{ISIC 2017 Challenge in Skin Lesion Analysis Towards Melanoma Detection}. We have participated in the \emph{Part 3: Lesion Classification} with a system for automatic diagnosis of nevus, melanoma and seborrheic keratosis. Our approach aims to incorporate the expert knowledge of dermatologists into the well known framework of Convolutional Neural Networks (CNN), which have shown impressive performance in many visual recognition tasks. In particular, we have designed several networks providing lesion area identification, lesion segmentation into structural patterns and final diagnosis of clinical cases. Furthermore, novel blocks for CNNs have been designed to integrate this information with the diagnosis processing pipeline.
\end{abstract}

\begin{figure}[ht]
  \centering
  \includegraphics[width=0.8\linewidth]{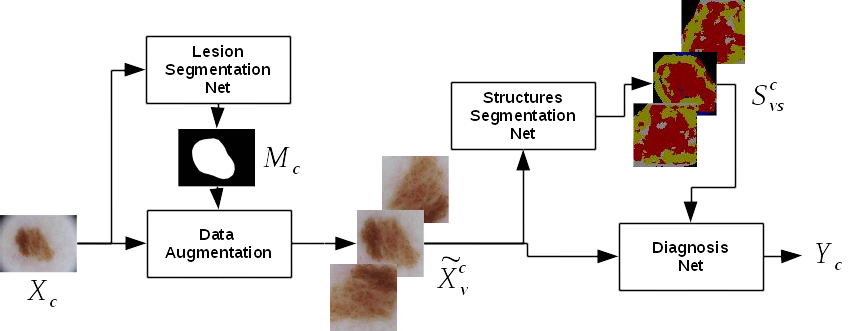}
  \caption{Main processing pipeline of our Automatic Diagnosis System \label{fig:main_pipeline}}
\end{figure}

\section{General description of the system}
The main pipeline of our system is depicted in Fig. \ref{fig:main_pipeline}. It comprises the following steps:

\begin{enumerate}
\item For each clinical case $c$, a dermoscopic image $X_c$ feeds a \emph{Lesion Segmentation Network} that generates a binary mask $M_c$ outlining the area of the image which corresponds to the lesion. The description of this module is given in section \ref{sec:LSN}.

\item Each clinical case $c$, which is now defined by the image-mask couple $\{X_c,M_c\}$, goes through the \emph{Data Augmentation Module}. This module aims to extend the initial visual support of the lesion by generating new views $v$ corresponding to different rotations and cropped areas. Hence, the output of this module is an extended set of images $\tilde{X}_v^c$ related to the lesion. Section \ref{sec:DAM} provides a detailed description of this data augmentation process.

\item The next step in the process is the \emph{Structure Segmentation Network}. It aims to segment each view of the lesion $\tilde{X}_v$ into a set of eight global and local structures that have turned to be very important for dermatologists in their daily diagnosis. Examples of these structures are dots/globules, regression areas, streaks, etc. Hence, the output of this system is a set of 8 segmentation maps $S_{vs}^c, s=1...8$, each one associated to a particular structure $s$ of interest. This module is introduced in section \ref{sec:SSN}.

\item Finally, the augmented set $\{\tilde{X}_v^c,S_{vs}^c\}$ is passed to the \emph{Diagnosis Network}, which is in charge of providing the final diagnosis $Y_c$ for the clinical case. The description of this network can be found in section \ref{sec:DN}.
\end{enumerate}

\section{Lesion Segmentation Network}
\label{sec:LSN}

The Lesion Segmentation Network has been developed by learning a Fully Convolutional Network (FCN) \citep{Shelhamer16}.  FCNs have achieved state-of-the-art results on the task of semantic image segmentation in general-content, as demonstrated in the PASCAL VOC Segmentation \citep{Everingham15}. In order to train a network for our particular task of lesion/skin segmentation, we have used the training set for the lesion segmentation task in the 2017 ISBI challenge. Let us note that the goal of this module is not to generate very accurate segmentation maps of a lesion, but to broadly identify the area of the image that corresponds to the lesion, giving place to a binary map $M_c$ for each clinical case.  

\begin{figure}[ht]
  \centering
  \includegraphics[width=0.1\linewidth]{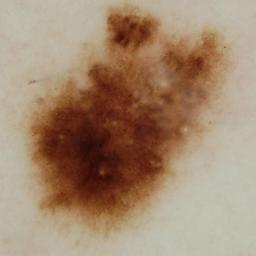}\hspace{0.1cm}\includegraphics[width=0.1\linewidth]{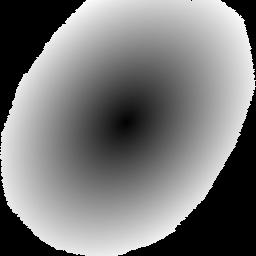}\hspace{0.1cm}\includegraphics[width=0.1\linewidth]{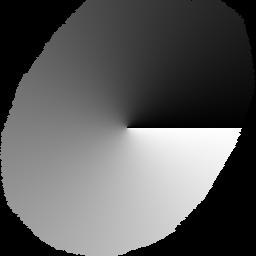}
  \caption{Example of a rotated and cropped view of a lesion and its Normalized Polar Coordinates. (Left) View of the lession (Middle) Normalize Ratio (Right) Angle \label{fig:NPC}}
\end{figure}

\section{Data Augmentation Module and Normalized Polar Coordinates}
\label{sec:DAM}
It is well known that data augmentation notably boosts the performance of deep neural networks, mainly when the amount of training data is limited. Among all the potential image variations and artifacts, invariance to orientation is probably the main requirement of our method, as dermatologists do not follow a specific protocol during the capture of a lesion. Other more complex geometric transformations such as affine or projective transforms are less interesting here as the dermatoscope is normally placed just over and orthogonally to the lesion surface. The particular process of data augmentation is described next:
\begin{enumerate}
\item First, starting from the pair $\{X_c,M_c\}$, we generate a set of rotated versions. 
\item As rotating an image without losing any visual information requires incorporating new areas which were not present in the original view, we find and crop the largest inner rectangle ensuring that all pixels belong to the original image.
\item Finally, as our sub subsequent CNNs (Structure Segmentation and Diagnosis) require square input images of 256x256 pixels, we finally perform various squared crops which are in turn re-sized to the required dimensions.
\end{enumerate}
Considering the aforementioned rotations and crops, for each given clinical case $c$, we generate an augmented set of 24 images, represented by a tensor $\tilde{X}^c_v \in \mathcal{R}^{256\times256\times3}$, with $v=1...24$.

In addition, for each generated view $\tilde{X}^c_v$, we compute the \emph{Normalized Polar Coordinates} from the lesion mask. The goal of this new alternative coordinates is to support subsequent processing blocks by providing invariance against shifts, rotations, changes in size and even irregular shapes of the lesions. To do so, we transform pixel Cartesian coordinates $(x_i,y_i)$ into normalized polar coordinates $(\rho_i,\theta_i)$, where $rho_i \in [0,1]$ and $\theta_i \in [0, 2\pi)$ stand for the normalized ratio and angle, respectively. The process to compute this transformation is as follows: first, the mask of the lesion is approximated by an ellipse with the same second-order moments. Then, we learn the affine matrix that transforms the ellipse into a normalized (unit ratio) circle centered at (0,0). Figure \ref{fig:NPC} shows an example of a rotated and cropped view of a lesion, and its corresponding normalized polar coordinates. 

\section{Structure Segmentation Network}
\label{sec:SSN}
The goal of this module is, given an input view of the lesion $\tilde{X}^c_v$, to provide a corresponding segmentation into a pre-defined set of textural patterns and local structures that are of special interest for dermatologists in their diagnosis. In particular, we have considered a set of eight structures: 1) \emph{dots, globules and cobblestone pattern}, 2) \emph{reticular patterns and pigmented networks}, 3) \emph{homogeneous areas}, 4) \emph{regression areas}, 5) \emph{blue-white veil}, 6) \emph{streaks}, 7) \emph{vascular structures} and 8) \emph{unspecific patterns}.

The main challenge to develop this module is the generation of a strongly-labeled training dataset, in which each image has an associated ground truth pixel-wise segmentation. This kind of annotation is often hard to obtain as it requires a huge effort of the dermatologists to manually outline the segmentations. Alternatively, providing weak image-level labels indicating only which structural patterns are present in each lesion is much easier for dermatologists and therefore becomes more realistic. Hence, following this latter approach, we asked dermatologists of a collaborating medical institution, the \emph{Hospital Doce de Octubre} in Madrid, to annotate the \emph{ISIC 2016 training dataset} with the presence or absence of the 8 considered structures. In particular, we asked them to provide one labels for each structure: 0 if the structure is not present, 1 if is locally present, 2 if it is present and large enough to be considered a global pattern in the lesion.

Given this weakly-annotated dataset, we have built our approach over the work of \citep{Pathak15}, where the authors introduced a novel constrained optimization for weakly-labeled segmentation using CNNs. The output of this network is a reduced version of the input image (64x64 in our case) where, for each pixel location $x_i$, a softmax is used to transform the net outputs $f_i(x_i;\theta)$ into probabilities as follows:
\begin{equation}
p_i(x_i|\theta)=\frac{1}{Z_i}exp(f_i(x_i|\theta))
\end{equation}
where $\theta$ represents the parameters of the CNN, and $Z_i=\sum_{s=1...8} exp(f_i(s|\theta))$ is the partition function at the location $i$. The presence or absence of a class, as well as, an estimate of its size in the image, lead to particular constraints over the probability $P_s = \sum_i p_i(s|\theta)$ accumulated over all pixel locations in the segmentation map:
\begin{itemize}
\item If a structure $s$ is not present in an image, the constraint acts as an upper bound over the accumulated probability $P_s$, which has to be nearly zero.
\item If a structure $s$ is local in an image, we impose a lower and upper bound on the accumulated probability $P_s$ in the image to control the total area of the structure in the lesion.
\item If a structure $s$ is global in an image, we impose a lower bound on the accumulated probability $P_s$ in the image to ensure a minimum area corresponding to the structure.
\end{itemize} 

In order to adapt this approach to our particular scenario, we have developed a set of modifications over the original approach, namely:
\begin{itemize}

\item We observed that using simple softmax function lead to situations in which many constraints over local structures were obeyed by assigning some residual probability to every location in the segmentation map. From our point of view, this is an undesired behavior, as one would rather expect a small set of pixels showing large probabilities of belonging to the structure of interest. To overcome this limitation, we have used a parametric softmax $p_i(x_i|\gamma,\theta)=\frac{1}{Z_i}exp(f_i(\gamma x_i|\theta))$. The parameter $\gamma$ is a soft-approximation towards the max function, and large values lead to scenarios in which each location shows high probability just for very reduced set of structures. In our case, we have used a value of $\gamma=20$.

\item We added a new constraint that helps to learn structures that appear in spatial locations of the lesion: e.g. streaks tend to appear in the borders of a lesion. For that end, we accumulate probabilities $P_s$ only in those locations that will likely contain the intended structure. At this point, we have defined these areas of interest over the Normalized Polar Coordinates described in section \ref{sec:DAM}, which are more adequate than the original Cartesian coordinates.

\end{itemize}

We have implemented this module taking the well-known \emph{vgg-vdd} \citep{Simonyan14} (the same network used as initialization for the lesion segmentation module), removing the top layers, and using the \emph{ISIC 2016 training dataset} and the described constrained optimization with weak annotations \citep{Pathak15}. The output of this module is, for each view $v$ of a clinical case $c$, a tensor $S_{v}^c \in \mathcal{R}^{64\times64\times8}$ that contains the 8 probability maps of the considered structures.

\begin{figure}[ht]
  \centering
  \includegraphics[width=.8\linewidth]{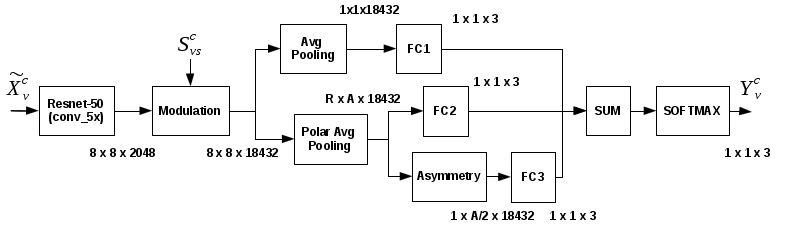}
  \caption{Processing pipeline of the Diagnosis Network \label{fig:diagnosis_pipeline}}
\end{figure}

\section{Diagnosis Network}
\label{sec:DN}
The \emph{Diagnosis Network} will gather the information from previous modules in order to generate a diagnose for each clinical case. As in the previous modules, our approach has taken a well-known CNN as starting point and modified the top layers to get a better adaptation to our problem.

The network chosen as basis is the \emph{resnet-50} \citep{He15}, which uses residual layers to avoid the degradation problem when more and more layers are stacked to the network. When applied to our 256x256 images, the last convolutional block (conv\_5x) of this network produces a tensor $T_c \in \mathcal{R}^{8 \times 8 \times 2048}$, which hopefully behaves as a detector of high level concepts (objects in Imagenet, the dataset for which it was originally designed). 

In the original work, an average pooling layer transformed this tensor into a single-value per channel and image $T_s \in \mathcal{R}^{1 \times 1 \times 2048}$, which was followed by a fully convolutional layer and a softmax to generate the final probabilities of the image containing the classes being detected. Hence, the goal of the average pooling was fusing detections at various locations of the input image and generating a unified score for each high-level concept. 

In our approach, however, we have modified the structure of the top layers of the network, giving place to the structure presented in Figure \ref{fig:diagnosis_pipeline}. We basically subdivide the top fully-connected layer providing the lesion diagnosis into three arms: a) the original arm with an average pooling followed by a fully connected layer (FC1), b) a second arm that performs a normalized polar pooling ($3x6$ rings by angles) and follows it by a fully connected layer (FC2), c) a third arm that estimates the asymmetry of lesion based on the previous polar pooling and applies then a Fully Connected layer (FC3). The results of the three arms are then linearly combined using a Sum block. We next describe the novel blocks that are required in this new structure and that have been specifically developed in this work:
\begin{enumerate}

\item \emph{Modulation block}: The goal of this block is to take advantage of the previous segmentations of the lesion into global and local structures which are of great interest for dermatologists in their daily diagnosis. To do so, this blocks fuses the previous structure segmentation maps $S_{v}^c$ with the filter outputs of the \emph{conv\_5x} layer in resnet-50. In particular, we modulate the outputs of the layer (2048 channels in our case) using the probabilities of the 8 local and global structures described in section \ref{sec:SSN}. By concatenating the resulting modulation with the original set of outputs we finally generate a set of channels which is 9 times the original one (18432 in our case). 

\item \emph{Polar Pooling}: This block aims to perform pooling operations over data (average or max pooling) but, rather than using rectangular spatial regions, we employ sectors defined in polar coordinates. Hence, this block is defined for a given number of radial rings R (radius ranging from 0 to 1) and angular sectors A (angles ranging between 0 and $2\pi$), producing an output of size $R \times A \times channels$. Furthermore, in order to adapt to the irregular shapes of the lesions, we use the normalized polar coordinates described in section \ref{sec:DAM}. Since, depending on the shape of the lesion and the size of the tensor being pooled, some combinations $(r,a)$ may not contain pixels in the image, we can also define overlaps between adjacent radius and angles to regularize the outputs. In addition, the division of the lesion into rings is non-uniform and ensures that every ring contains the same number of pixels for a perfect circular lesion.

\item \emph{Asymmetry}: This block computes metrics that evaluate the asymmetry of a lesion for a given angle. In particular, given a polar division of the lesion into $R\times A$ sectors, we compute the asymmetry for $A/2$ angles by folding the lesion over each angle and computing the accumulated absolute difference between corresponding sectors.  

\end{enumerate}

As shown in the Figure \ref{fig:diagnosis_pipeline}, we combine these modules to generate a final output $Y_v^c$ for each considered view of a clinical case. 

Finally, in order to generate a final output for each clinical case $Y_c$, we consider independence between views leading to a factorization:
\begin{equation}
	Y_c=\prod_{v=1}^V Y_v^c
\end{equation}

It is also worth noting that our final submission has also incorporated in the factorization an extra classifier which depends only on external information about the clinical case, such as patient gender and age, and lesion area.

\section{Code}
\label{sec:CODE}
The code that implements this paper as well as the Lesion Segmentation and Diagnosis Networks are provided in the following link:  \url{https://github.com/igondia/matconvnet-dermoscopy}.

\section*{Acknowledgments}
We kindly thank dermatologists of \emph{Hospital 12 de Octubre of Madrid} because of their inestimable help annotating the data contents with the weak labels of structural patterns. This work was supported in part by the National Grant TEC2014-53390-P and National Grant TEC2014-61729-EXP of the Spanish Ministry of Economy and Competitiveness. In addition, we gratefully acknowledge the support of NVIDIA Corporation with the donation of the TITAN X GPU used for this research.

\bibliographystyle{abbrvnat}
\bibliography{melanomas}

\end{document}